\documentclass[final]{cvpr}

\usepackage{times}
\usepackage{epsfig}
\usepackage{graphicx}
\usepackage{amsmath}
\usepackage{amssymb}

\usepackage{subcaption}
\usepackage{booktabs}% http://ctan.org/pkg/booktabs
\usepackage{gensymb}
\usepackage{siunitx}
\usepackage[dvipsnames, table]{xcolor}
\definecolor{superlightgray}{RGB}{232,232,232}
\usepackage{footmisc}
\usepackage{makecell}
\usepackage{multirow}

\usepackage[pagebackref=true,breaklinks=true,colorlinks,bookmarks=false]{hyperref}

\def\cvprPaperID{****} % *** Enter the CVPR Paper ID here

\def\cvprPaperID{****} % *** Enter the CVPR Paper ID here
\def\confYear{CVPR 2021}

\begin{document}

\title{TIMo -- A Dataset for Indoor Building Monitoring with a Time-of-Flight Camera}
%\author{
%\begin{tabular}{lcccr}
%Pascal Schneider&&Yuriy Anisimov&&Bruno Mirbach\\
%Raisul Islam&&Jason Rambach&&Didier Stricker
%\end{tabular}
%\\\\
%DFKI -- German Research Center for Artificial Intelligence\\
%Trippstadter Str. 122, 67663 Kaiserslautern, Germany\\
%{\tt\small firstname.lastname@dfki.de

\author{Pascal Schneider\textsuperscript{*}, Yuriy    Anisimov\textsuperscript{*}, Raisul Islam\textsuperscript{*}, Bruno Mirbach\textsuperscript{*},
\\ Jason Rambach\textsuperscript{*}, Frédéric Grandidier\textsuperscript{\textdagger} and Didier Stricker\textsuperscript{*}
\\\\
* DFKI -- German Research Center for Artificial Intelligence, 
{\tt firstname.lastname@dfki.de}\\
\textdagger \hspace{2pt} IEE S.A., Luxembourg, {\tt frederic.grandidier@iee.lu}
}

\maketitle

\begin{abstract}
	We present \emph{TIMo} (\textbf{T}ime-of-flight \textbf{I}ndoor \textbf{Mo}nitoring), a dataset for video-based monitoring of indoor spaces captured using a time-of-flight (ToF) camera. The resulting depth videos feature people performing a set of different predefined actions, for which we provide detailed annotations. Person detection for people counting and anomaly detection are the two targeted applications. Most existing surveillance video datasets provide either grayscale or RGB videos. Depth information, on the other hand, is still a rarity in this class of datasets in spite of being popular and much more common in other research fields within computer vision. Our dataset addresses this gap in the landscape of surveillance video datasets. The recordings took place at two different locations with the ToF camera set up either in a top-down or a tilted perspective on the scene. The dataset is publicly available at \url{https://vizta-tof.kl.dfki.de/timo-dataset-overview/}.
\end{abstract}

\begin{figure}[t]
	\begin{center}
		%\fbox{\rule{0pt}{1in} \rule{0.9\linewidth}{0pt}}
		\begin{subfigure}[b]{0.225\textwidth}
			\centering
			\includegraphics[width=38mm]{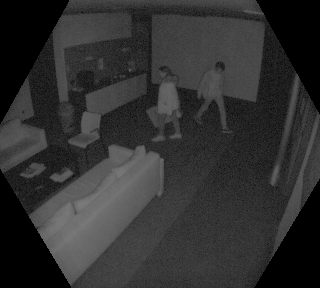}
		\end{subfigure}
		\begin{subfigure}[b]{0.225\textwidth}
			\centering
			\includegraphics[width=38mm]{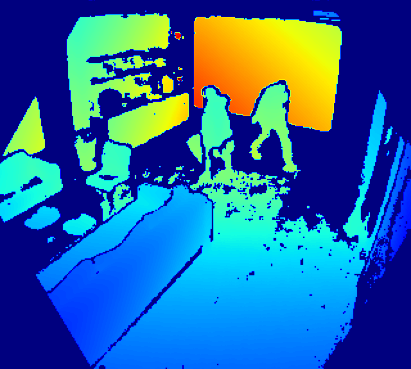}
		\end{subfigure}
		\\
		\begin{subfigure}[b]{0.225\textwidth}
			\centering
			\includegraphics[width=38mm]{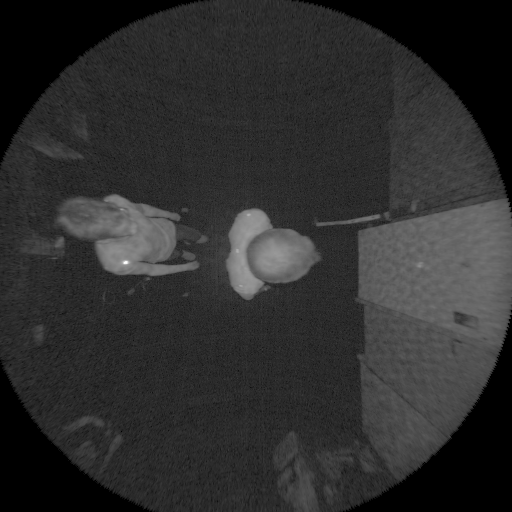}
		\end{subfigure}
		\begin{subfigure}[b]{0.225\textwidth}
			\centering
			\includegraphics[width=38mm]{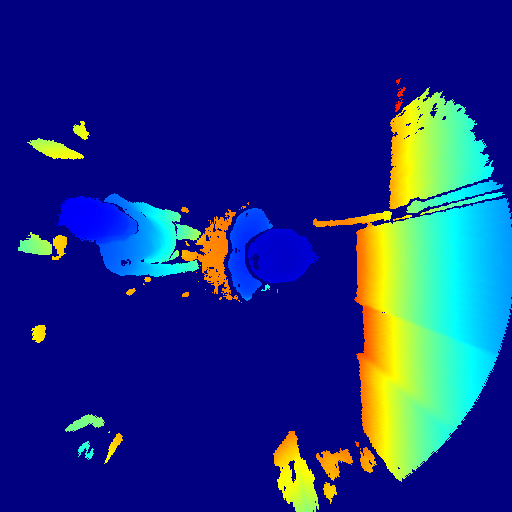}
		\end{subfigure}
	\end{center}
	\caption{Example frames from our dataset. Top: a scene from tilted view, Bottom: a scene from top-down view. Left: IR image: Right depth image with depth encoded as color}
	\label{fig:example_images}
\end{figure}

\section{Introduction}
Traditionally, surveillance cameras are RGB or IR cameras.
For the realization of robust automatic building management functions, time-of-flight depth cameras offer, however, some unique benefits. First of all they are more robust to illumination and color variations, and allow natural geometrical background removal.
The depth information they provide allows to detect, classify and localize persons and objects precisely in 3D space.
Moreover people are much less likely to be identifiable in depth data compared to RGB. 
Thus, monitoring places and at the same time preserving people's privacy can be reconciled to a much better degree.
For these reasons, the first building management systems based on time-of-flight technology have have been around for about 10 years. (see \footnote{\url{https://www.iee-sensing.com/en/building-management-security.html}}), which realize a small set of building management functions such as people counting or single access control. 
Future time-of-flight sensors with higher sensitivity and resolution in combination with novel deep-learning algorithms promise to both enhance the performance of existing building management systems and to realize novel functions as the behaviour analysis of persons and detection of anomalous situations. 

The increasing affordability of consumer depth cameras has helped the analysis of depth data from time-of-flight sensors to become a popular and established part of current computer vision research. Examples of frequently used devices in this context are the \emph{Microsoft Kinect} \cite{Zhang12}, \emph{Intel RealSense} \cite{KeselmanWGB17} or \emph{Asus Xtion} \cite{gonzalez2013metrological}. The benefits of depth images to provide 3D geometric information has made it a widely used data modality for certain applications such as action recognition \cite{survey_rgbd_action_recognition} or gesture classification \cite{wang2016large}. With our dataset we want to foster research towards robust and high performing person detection and novel smart building functions such as the detection of anomalies.

An advantage of our dataset over existing ones is the choice of more modern hardware. Our recordings were captured using a \emph{Microsoft Azure Kinect} camera, which features higher image resolution, higher field-of-view and lower distance error compared to the \emph{Microsoft Kinect v2}. \cite{tolgyessy2021evaluation, WasenmullerS16}.

Some anomaly datasets additionally suffer from a very limited number of anomalous events. This is sometimes due to the fact that they use actual surveillance footage where anomalies tend to be rare. While there are advantages to using data from real-world anomalies, it can severely limit the amount of data available for testing anomaly detection algorithms. Our dataset features a large number of anomalies that facilitates testing at scale.

The rise of deep neural networks as the primary approach for solving computer vision problems leads to a demand for very large datasets that support training and testing complex models. Our dataset consists of about \num{1500} recordings with \num{44} different subjects and in total sums up to more than \num{612000} individual video frames, which makes it suitable for the development of data-hungry approaches. Additionally, \num{243} sequences with \num{22700} labeled frames with 3D bounding boxes and segmentation mask are provided for the person detection with counting purposes. It is moreover relatively easy to generate high quality synthetic data for depth videos in case the real data is not sufficient, since computation of depth is a customary step of computer graphics pipelines anyway. The \emph{TiCAM} dataset is an example of how synthetic depth data can complement real data \cite{katrolia2021ticam}.

The rest of the paper is structured as follows: \secref{sec:rel_work} puts our dataset into context w.r.t.\ existing ones. \secref{sec:dataset} then details the content and process of acquisition of our recordings and the accompanying annotations.

\section{Related Work} \label{sec:rel_work}
The popularity of depth-sensing technology led to a large number of datasets to be released featuring RGB videos with an additional depth channel, or, in short, RGB-D. While pure RGB datasets can sometimes be compiled using existing recordings (such as videos on online video platforms), RGB-D data usually has to be recorded specifically for the dataset. A review of some of these datasets is given in \cite{rgbd_datasets_2017}. Some provide static scans of indoor spaces, e.g.\ \emph{Stanford 2D-3D-Semantics} \cite{s3dis_dataset} or \emph{ScanNet} \cite{scannet_dataset} and are commonly used for tasks such as semantic segmentation of point clouds. 

Our dataset differs from this type of dataset in that it does not focus on the reconstruction of the static geometry of the indoor space but instead on capturing human action performed within the scene. In this regard it is more similar to datasets used for action recognition such as \emph{UTD-MHAD} \cite{utd_mhad_dataset} or \emph{NTU RGB+D} \cite{ntu_rgbd} and \emph{NTU RGB+D 120} \cite{ntu_rgbd_120_dataset}. However, these datasets were not created for the applications of anomaly detection or people counting. The camera angles and the nature of the recorded scenes make them unsuitable for these tasks. Therefore, there is a need for datasets particularly geared towards the development of algorithms for monitoring indoor spaces.

Existing datasets for anomaly detection usually provide only RGB data, e.g.\ the \emph{Shanghai-Tech} dataset \cite{shanghaitech_dataset} or \emph{UCF-Crime} \cite{ucf_crime_dataset}. Our dataset fills this existing gap by providing depth data of realistic scenarios and following camera angles as they would be common in a surveillance context.

Unlike some other datasets  -- such as \cite{ntu_rgbd_120_dataset} -- we did not aim for a large variety of backgrounds or illumination in the data. This limitation is a consequence of the time-consuming process of calibration and the risk of having correlations between the background and the content of sequences, which has the potential to compromise the learning process and results when using machine learning. In addition, background variations are less relevant in depth data. We therefore committed to record only a few scenes and put more focus on a large and well balanced variance of subjects and actions and on providing high quality supplementary information, e.g.\ camera calibration parameters.

\begin{table*}[ht]
\centering
	\begin{tabular}{| c || c | c | c | c | c | c |} 
		\hline
		\textbf{Dataset} & \textbf{Year} & \makecell{\textbf{\# Sequences} \\ \textbf{(\# Frames)}} & \makecell{\textbf{Data} \\ \textbf{Modalities}} & \makecell{\textbf{Camera} \\ \textbf{Hardware}} & \textbf{Annotations} & \makecell{\textbf{Environ-}\\\textbf{ment}}\\ [0.5ex] 
		\hline\hline
		\makecell{\textbf{TIMo Anomaly} \\ \textbf{Detection} (ours)}    & 2021 & \makecell{1588 \\ (612K)}& IR, Depth & \makecell{MS Kinect \\ Azure} & \makecell{Anomaly Frames} & Indoor\\
		\hline
		\makecell{\textbf{TIMo Person} \\ \textbf{Detection} (ours)}    & 2021 & \makecell{243 \\ (23.6K)} & IR, Depth & \makecell{MS Kinect \\ Azure} & \makecell{2D + 3D Object BBox, \\ 2D Segm. Masks} & Indoor\\
		\hline
		\makecell{\textbf{ShanghaiTech} \\ \textbf{Campus} \cite{shanghaitech_dataset}} & 2018 & \makecell{437 \\ (317K)}& RGB & \makecell{RGB \\ Camera} & \makecell{Anomaly Frames, \\ Anomaly Masks} & Outdoor\\ 
		\hline
	    \makecell{\textbf{UTD-MHAD} \\ \cite{utd_mhad_dataset}}    & 2015 &  \makecell{861 \\ (45K)}& \makecell{RGB, Depth,\\3D Joints, ID } & MS Kinect v1 & \makecell{Action Classes} & Indoor\\
		\hline
		\makecell{\textbf{NTU-RGB+D} \\ \textbf{120} \cite{ntu_rgbd_120_dataset}}    & 2019 & \makecell{114K \\ (4M)} & \makecell{RGB, Depth, \\ 3D Joints, Inertia} & MS Kinect v2 & \makecell{Action Classes} & Indoor\\
		\hline
		\makecell{\textbf{UCF-Crime} \\ \cite{ucf_crime_dataset}}    & 2018 & \makecell{1900 \\ (13.8M)} & RGB & \makecell{RGB \\ Camera} & \makecell{Anomaly Frames} & \makecell{Indoor + \\ Outdoor}\\
		\hline
		\makecell{\textbf{TiCAM} \\ (Real) \cite{katrolia2021ticam}}    & 2021 & \makecell{\makecell{533} \\ (6.7K \\ / 118K)} & \makecell{RGB, IR, \\ Depth} & \makecell{MS Kinect \\ Azure} & \makecell{2D + 3D Object BBox, \\ 2D Segm. Masks \\ / Action Classes} & \makecell{Car Cabin}\\
		\hline
		\makecell{\textbf{DAD} \\ \cite{dad_dataset}}    & 2020 & \makecell{386 \\ (2.1M)} & IR, Depth & \makecell{CamBoard \\ pico flexx} & \makecell{Anomaly Frames} & \makecell{Car Cabin}\\
		\hline
		\makecell{\textbf{CUHK} \\ \textbf{Avenue} \cite{cuhk_ave_dataset}}    & 2013 & \makecell{37 \\ (31K)} & RGB & \makecell{RGB \\ Camera} & \makecell{Anomaly Frames, \\ Anomaly BBoxes} & \makecell{Outdoor}\\
		\hline
		%\makecell{\textbf{UMN} \\ \cite{?}}    & 2006 & 7.7K & 5\makecell{??? \\ (???)} & ???? & \makecell{RGB \\ Camera}  & ???? & ??\\
		%\hline
     	\makecell{\textbf{UCSD} \\ \textbf{Ped 1 + 2} \cite{ucsd_ped_dataset}}    & 2010 &\makecell{70 + 28 \\ (14K + 4.6K)} & Greyscale & \makecell{Greyscale \\ Camera} & \makecell{Anomaly Frames, \\ Anomaly Masks\footnote{Masks are only provided for a subset of the dataset.}} & Outdoor\\
		\hline
		\makecell{\textbf{Subway Exit} \\ \textbf{+ Entrance} \cite{subway_dataset}}    & 2008 & \makecell{1 + 1 \\ (137K + 72K)} & Greyscale & \makecell{Greyscale \\ Camera} & \makecell{Anomaly Frames,\\ rough Anomaly Location} & \makecell{Subway \\ Station}\\
		\hline
		\makecell{\textbf{IITB-Corridor}\\ \cite{iitb_dataset}}    & 2020 & \makecell{368 \\ (484K)} & RGB & \makecell{RGB \\ Camera} & \makecell{Anomaly Frames} & \makecell{Outdoor \\ (Corridor)}\\
		\hline
	\end{tabular}
	\caption{Comparison of related datasets to our dataset. 3D joints refers to joints of the human body such as the are used in pose estimation.}
	\label{tab:dataset_comparison}
\end{table*}

\section{TIMo Dataset} \label{sec:dataset}
We describe the content of the dataset as well as the process of recording. \secref{sec:setup} and \ref{sec:acquisition} cover the choice of hardware and scene and how recordings were carried out. \secref{sec:postprocessing} -- \ref{sec:annotations} give more details about the content and annotations we provide in the dataset.

\subsection{Data Modalities} \label{sec:data_modalities}
We provide the infrared (IR) images and the depth maps estimated by the Microsoft Azure Kinect camera. The camera in principle also features recording RGB images, which we use in some of the figures for visualization purposes. Note, however, that our focus lies on depth data and hence we do not provide RGB frames in the public dataset and there are also no annotations for the RGB modality.

\subsection{Setup}\label{sec:setup}
Recordings took place at two different locations, which we will refer to as \emph{Scene 1} and \emph{Scene 2}. 
\emph{Scene 1} is an open office area with a small kitchen and a seating area. For this scene the Microsoft Azure Kinect camera was installed in two different positions.
There was a tilted-view mounting in which the camera was able to monitor a large portion of the room including the four entrance possibilities, as shown in \figref{fig:dfki_entrances}. In addition, the camera was mounted on a metal frame above the entrance {\em B} with a top-down view, monitoring people entering or leaving the room through a hallway. 

\begin{figure*}[t]
	\begin{center}
		\begin{subfigure}[]{0.32\textwidth}
			\centering
			\includegraphics[width=\textwidth]{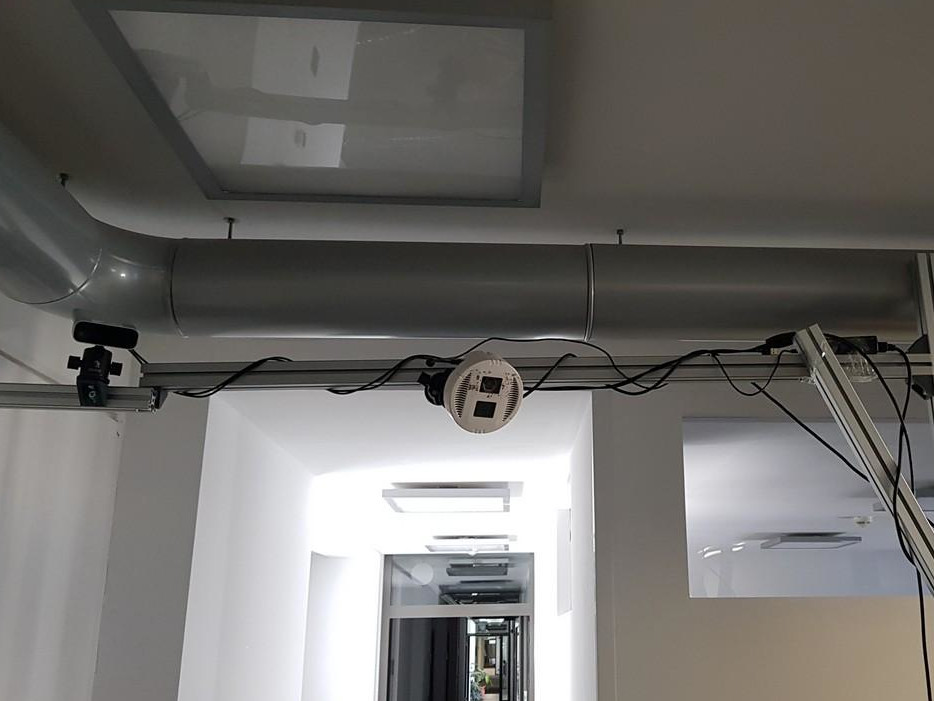}
			\caption{Camera mounting for tilted-view scene.}
		\end{subfigure}
		\begin{subfigure}[]{0.32\textwidth}
			\centering
			\includegraphics[width=\textwidth]{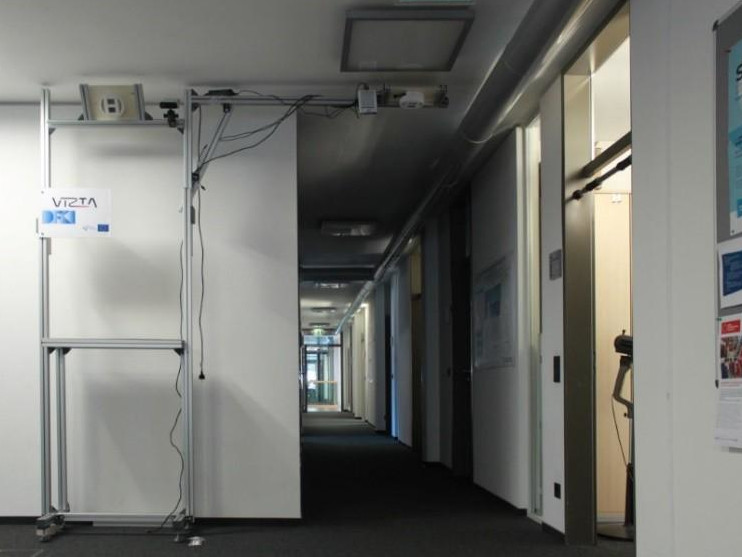}
			\caption{Setup for top-down view in \emph{Scene 1}.}
		\end{subfigure}
		\begin{subfigure}[]{0.32\textwidth}
			\centering
			\includegraphics[width=\textwidth]{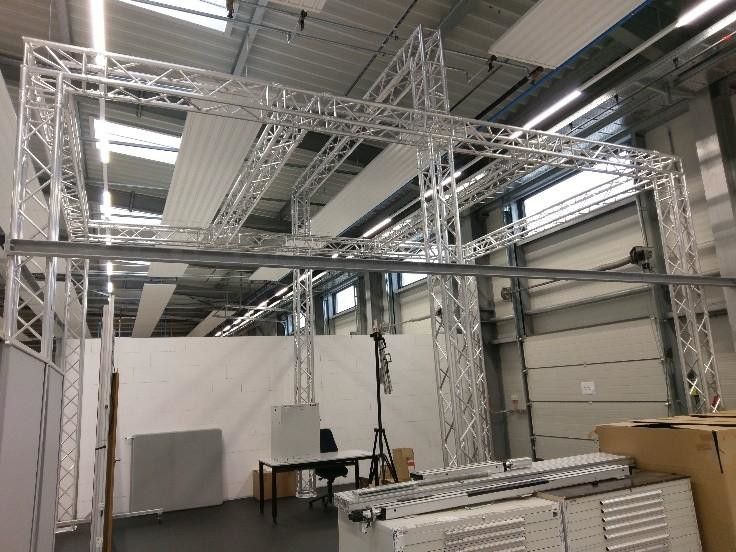}
			\caption{Setup for top-down view in \emph{Scene 2}.}
		\end{subfigure}
	\end{center}
	\caption{Recording setups used for capturing the dataset.}
	\label{fig:recording_setup}
\end{figure*}

The same top-down mounting orientation has been used in {\em Scene 2} but with more flexibility. There the monitored area is less confined, allowing persons to cross the scene in all directions. Moreover, the camera is mounted on a lift allowing to vary the camera height. \figref{fig:recording_setup} shows the different
camera mounting setups. For the two different mounting orientations, the camera configuration was different as described below. 

\subsubsection{Top-down View} \label{sec:top_down_view}
For top-down recordings we used the wide field-of-view (WFOV) configuration of the Azure Kinect (\SI{120}{\degree}$\,\times\,$\SI{120}{\degree}). The native resolution of the sensor in WFOV configuration is \num{1024}$\,\times\,$\num{1024} pixels. We employ a \num{2}$\,\times\,$\num{2} binning technique which reduces the resolution down to \num{512}$\,\times\,$\num{512} pixels but at the same time yields a higher operating range, which is \SI{0.25}{\m} -- \SI{2.88}{\m}. The capturing rate is set to \num{30} frames per second (FPS).
The camera height above ground was varied between \SI{2.25}{m}, \SI{2.50}{m} and \SI{2.75}{m}

\subsubsection{Tilted-view} \label{sec:tilted_view}
For recordings from the tilted view we used the narrow field-of-view (NFOV) configuration (\SI{75}{\degree}$\,\times\,$\SI{65}{\degree}). The native resolution of the sensor in NFOV configuration is \num{576}$\,\times\,$\num{640} pixels. Same as for the top-down configuration, we use a frame rate of \num{30} FPS and \num{2}$\,\times\,$\num{2} binning, which results in a resolution of \num{288}$\,\times\,$\num{320} pixels.

The maximum guaranteed depth operating range for the NFOV configuration with \num{2}$\,\times\,$\num{2} binning is \SI{5.46}{\m}. For the recordings from the tilted-view camera this is not enough range to cover the complete scene. As is usual for time-of-flight cameras, the data quality also depends significantly on the remission properties of the surface in question. Nevertheless we observed that we get adequate depth measurements for the relevant parts of the scene up to about \SI{10}{\m} in our setup. 

\subsection{Acquisition} \label{sec:acquisition}
\begin{figure}[b]
	\centering
	\includegraphics[width=0.47\textwidth]{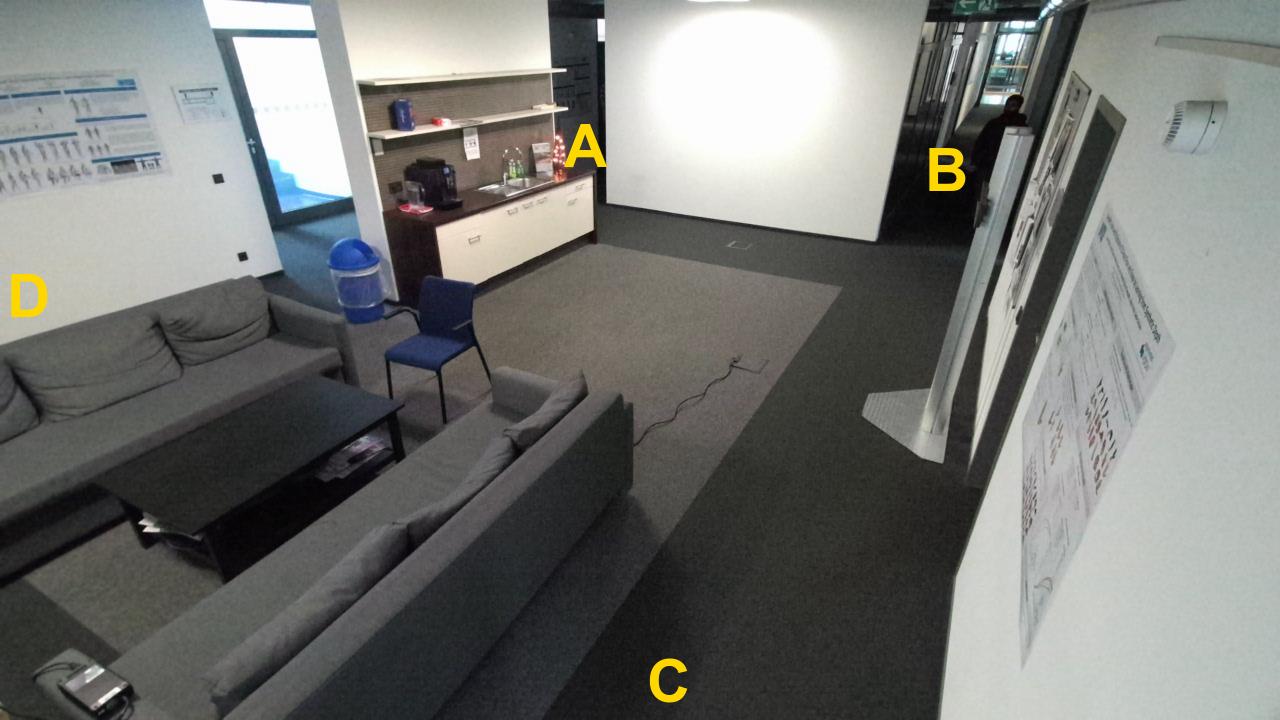}
	\caption{Tilted-view scene with marked entrances.}
	\label{fig:dfki_entrances}
\end{figure}

The test cases were defined prior to recording according to a test matrix, which aims at preventing unintentionally introduced correlations in the dataset. The anomalies in the anomaly dataset also belong to pre-defined test cases.
The camera was calibrated according to a world coordinate system before each recording session. 
For recording, the test subjects were instructed to enter and leave the scene through a specific entrance. For the tilted-view scene there are four such entrances (see \figref{fig:dfki_entrances}). In the top-down-view scenes the test subjects cross the scene either in the $X-$ or $Y-$ direction of the camera coordinate system. The subjects perform a given action after entering the scene and then leave again. The instructions on how an action was to be performed were kept rather vague in order to have some degree of variance between performances of the same action type. A full list of the choreographies can be found on the dataset website\footnote{\label{data_format_website}\url{https://vizta-tof.kl.dfki.de/building-data-format/}}.

\subsection{Post-Processing} \label{sec:postprocessing}
Post-processing of the data and the annotations is kept to a minimum in order to allow users to choose between using the raw data or applying custom normalization themselves. Images in the person detection dataset are undistorted and remapped to the common pinhole camera model. The original rotation and translation matrices are also provided per sequence for its further conversion to the 3D world coordinate system (e.g. in a form of a point cloud). The segmentation masks and 2D bounding boxes are provided in the coordinates of the undistorted and remapped images (see \secref{sec:annotations} for more details on the annotations).

\subsection{Data Format} \label{sec:dataformat}
The IR and depth videos are stored as individual frames in the \emph{Portable Network Graphics} (PNG) format with a single 16-bit channel. The pixel values in the depth images directly correspond to the depth measurements in millimeters. A pixel value of \num{0} is used as a special value to indicate that there is no valid depth estimation for this pixel. Note that the example frames shown in this paper have been transformed w.r.t.\ value range and contrast for better visualization.

The video sequences and individual frames follow a common naming scheme which includes the most relevant information directly in the file name. It includes the choreography, a sequence ID, the camera height, a timestamp and a calibration ID. More details on the naming scheme can also be found on the website \footref{data_format_website}.

\subsection{Annotations} \label{sec:annotations}
\begin{figure*}[ht]
	\begin{center}
		\begin{subfigure}[]{0.62\textwidth}
			\centering
			\includegraphics[width=\textwidth]{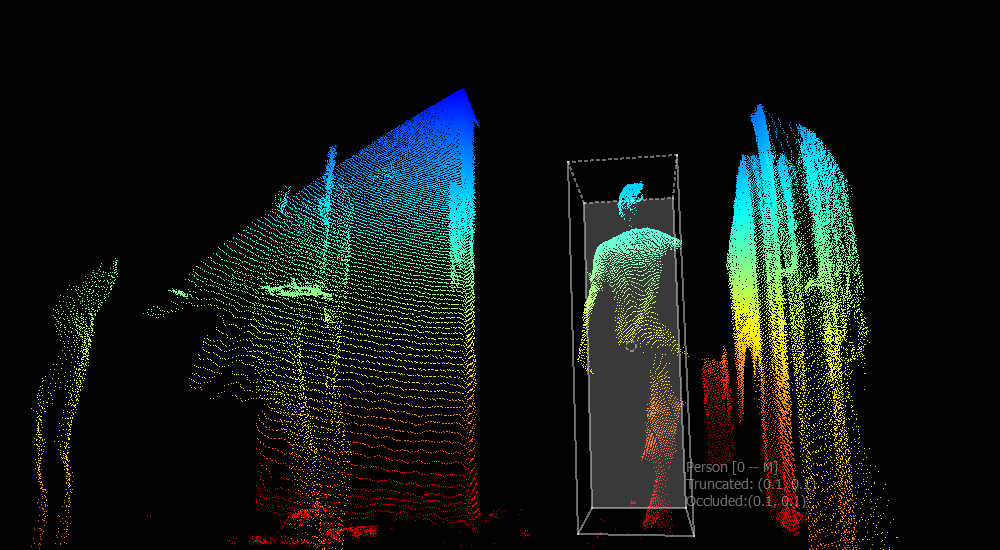}
			\caption{Example of bounding box in the 3D domain.}
		\end{subfigure}
		\begin{subfigure}[]{0.343\textwidth}
			\centering
			\includegraphics[width=\textwidth]{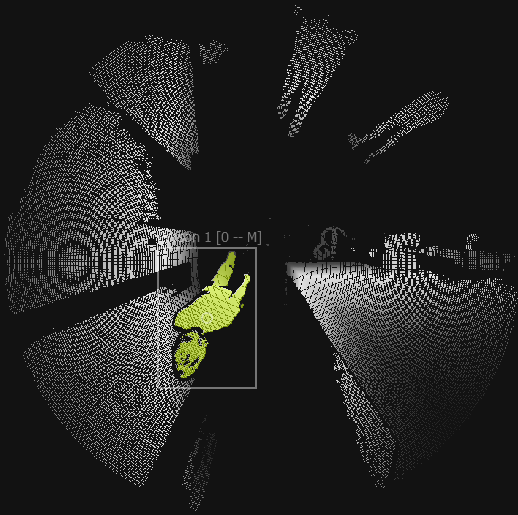}
			\caption{Example of 2D bounding box.}
		\end{subfigure}
	\end{center}
	\caption{Visualization of annotations for the person detection/people counting dataset, generated by \cite{stumpf2021salt}.}
	\label{fig:example_annotations}
\end{figure*}

We provide annotations for people and objects in the form of 2D and 3D bounding boxes and segmentation masks for the person detection dataset and anomaly annotations for the anomaly dataset. 
For the anomaly dataset both tilted view data recorded in {\em Scene 1} and top-down data recorded in {\em Scene 2} were annotated. For the person detection dataset data from both scenes recorded in the top-down configuration were annotated.
The annotations were done manually using a software tool that was developed specifically for the purpose of annotating 3D data \cite{stumpf2021salt}.

\subsubsection{Anomaly Annotations} \label{sec:anomaly_annotations}
\begin{figure*}[t]
	\begin{center}
		%\fbox{\rule{0pt}{1in} \rule{0.9\linewidth}{0pt}}
		\begin{subfigure}[]{0.245\textwidth}
			\centering
			\includegraphics[width=\textwidth]{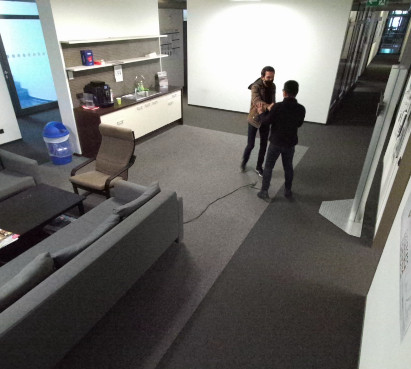}
		\end{subfigure}
		\begin{subfigure}[]{0.245\textwidth}
			\centering
			\includegraphics[width=\textwidth]{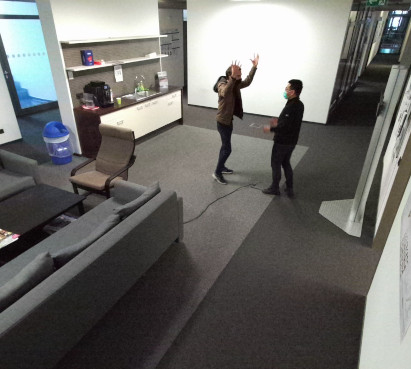}
		\end{subfigure}
		\begin{subfigure}[]{0.245\textwidth}
			\centering
			\includegraphics[width=\textwidth]{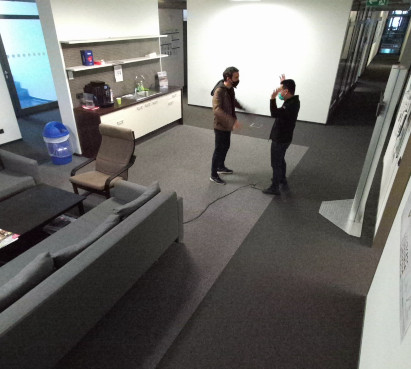}
		\end{subfigure}
		\begin{subfigure}[]{0.245\textwidth}
			\centering
			\includegraphics[width=\textwidth]{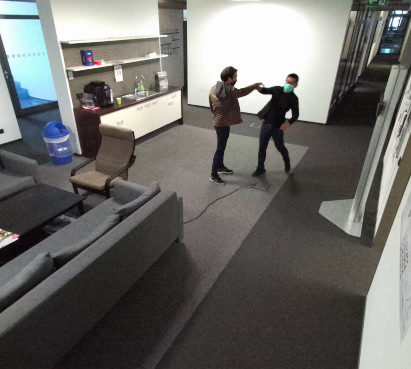}
		\end{subfigure}
		\\
		\begin{subfigure}[]{0.245\textwidth}
			\centering
			\includegraphics[width=\textwidth]{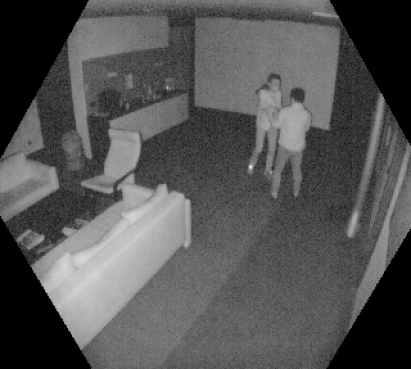}
		\end{subfigure}
		\begin{subfigure}[]{0.245\textwidth}
			\centering
			\includegraphics[width=\textwidth]{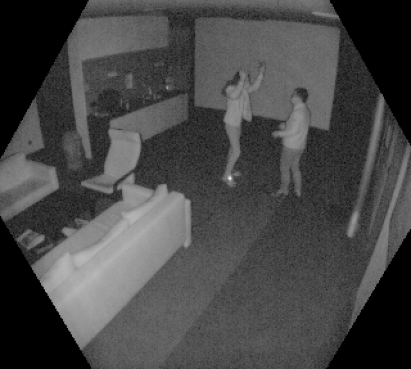}
		\end{subfigure}
		\begin{subfigure}[]{0.245\textwidth}
			\centering
			\includegraphics[width=\textwidth]{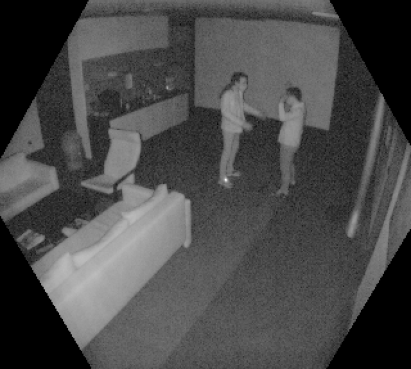}
		\end{subfigure}
		\begin{subfigure}[]{0.245\textwidth}
			\centering
			\includegraphics[width=\textwidth]{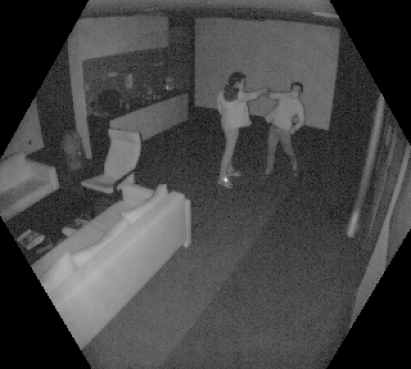}
		\end{subfigure}
		\\ \vspace{1em}
		\begin{subfigure}[]{0.245\textwidth}
			\centering
			\includegraphics[width=\textwidth]{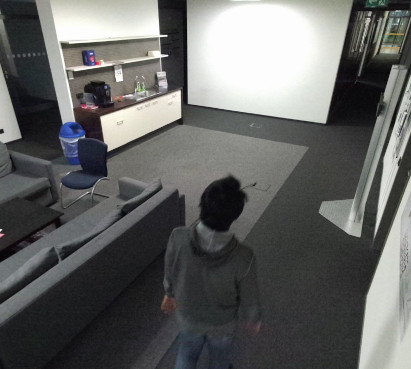}
		\end{subfigure}
		\begin{subfigure}[]{0.245\textwidth}
			\centering
			\includegraphics[width=\textwidth]{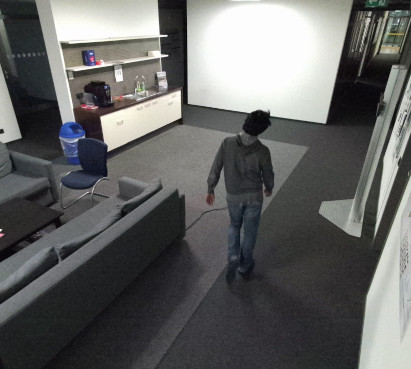}
		\end{subfigure}
		\begin{subfigure}[]{0.245\textwidth}
			\centering
			\includegraphics[width=\textwidth]{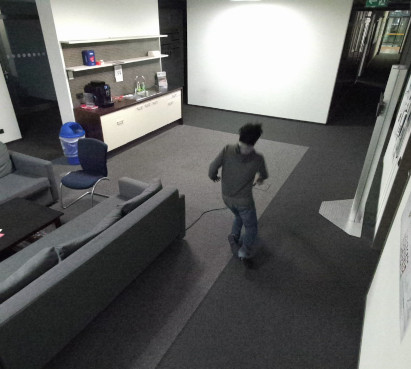}
		\end{subfigure}
		\begin{subfigure}[]{0.245\textwidth}
			\centering
			\includegraphics[width=\textwidth]{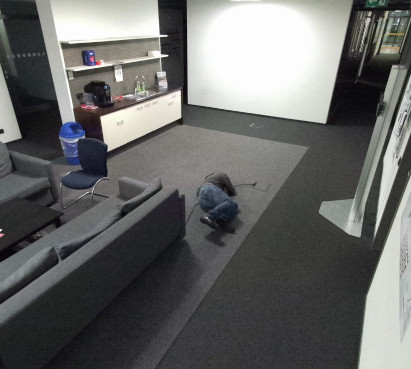}
		\end{subfigure}
		\\
		\begin{subfigure}[]{0.245\textwidth}
			\centering
			\includegraphics[width=\textwidth]{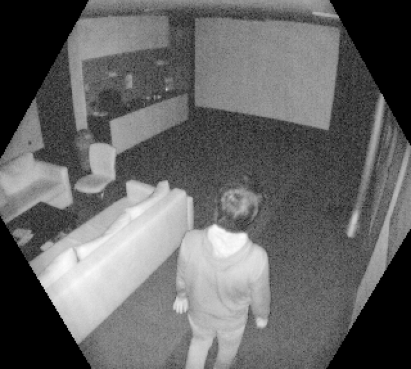}
		\end{subfigure}
		\begin{subfigure}[]{0.245\textwidth}
			\centering
			\includegraphics[width=\textwidth]{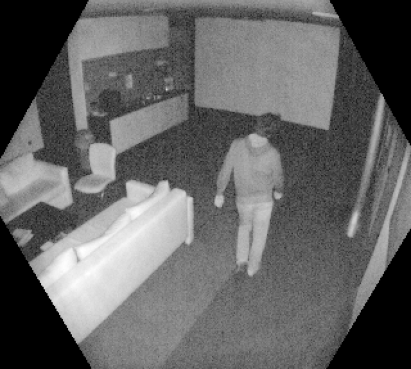}
		\end{subfigure}
		\begin{subfigure}[]{0.245\textwidth}
			\centering
			\includegraphics[width=\textwidth]{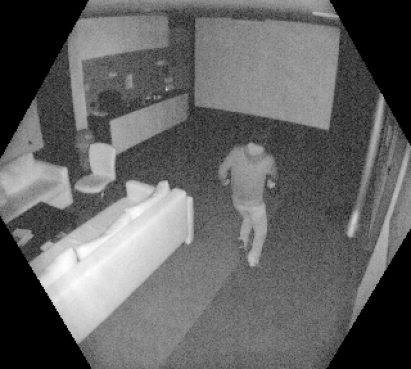}
		\end{subfigure}
		\begin{subfigure}[]{0.245\textwidth}
			\centering
			\includegraphics[width=\textwidth]{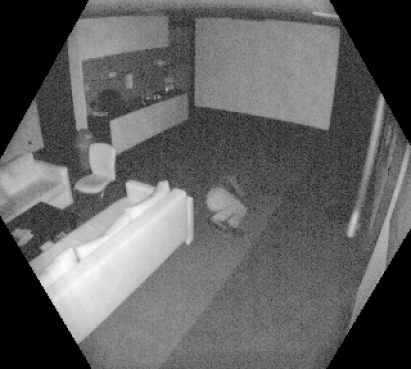}
		\end{subfigure}
	\end{center}
	\caption{Examples of anomalies. The top row shows frames from a sequence with an argument (\texttt{CROSS\_ARG}) between two people, the bottom row a collapsing person (\texttt{COLLAPSE}).}
	\label{fig:example_anomalies_IR}
\end{figure*}

Anomaly annotations are provided as pairs of frame indices which indicate when the anomalous event within the given sequence starts and ends. Note that the frames at \texttt{START\_FRAME\_IDX} and \texttt{END\_FRAME\_IDX} both also belong to the anomalous event, thus making it $\texttt{END\_FRAME\_IDX} - \texttt{START\_FRAME\_IDX} + 1$ frames long.

Examples of anomalies include left behind objects, people arguing or throwing objects. \figref{fig:example_anomalies_IR} illustrates two instances of anomalous events within the dataset.

All choreographies that are not labeled as anomalous are consequently considered to be normal. This includes activities such as getting coffee from the kitchen, talking with one another or simply walking.

\subsubsection{Person and Object Annotations} \label{sec:person_and_object_annotations}
For the person detection we provide annotations for people and objects in the following form: 

\begin{itemize}
	\item 2D segmentation masks per frame, saved as 8-bit PNG images. Pixel values correspond to class and instance IDs respectively;
	\item 2D bounding boxes per annotation, described as pixel coordinates of rectangular around the annotated object. This data is presented in corresponding CSV in a form of $[x_1, y_1, x_2, y_2]$, where $(x_1, y_1)$ and $(x_2, y_2)$ are the start and end coordinates of the bounding box;
	\item 3D bounding boxes per annotation, presented in corresponding CSV as a box center $(cx, cy, cz)$ and its dimensions $(dx, dy, dz)$ in the world coordinate system.
\end{itemize}

Objects are annotated as a separated class only if they are not held by a person at the specific frame. The 3D bounding boxes are generated automatically using the segmentation masks, since the calibration of the camera allows a direct mapping between the 2D image space and the 3D point clouds implicitly given in the form of the depth maps. We also used interpolation between manual annotations to speed up the process in situations where this could be done without impairing the quality of the annotations. All annotations were additionally validated by a person different from the one who created the annotation.

Examples of these annotations are illustrated in \figref{fig:example_annotations}.
\begin{table}[h]
\centering
\begin{tabular}{|r||ccc|}
\hline
\rowcolor{superlightgray} \multicolumn{4}{|c|}{\textbf{Person Detection Dataset}} \\
\hline
\rowcolor{superlightgray} \textbf{Data Type}        & \textbf{Sequences} & \textbf{Frames} & \makecell{\textbf{Anno-} \\\textbf{tations}} \\
\hline \hline 
\textbf{Training}                  & 125                & 6,415            & 8,501                 \\
\hline
\textbf{Complex Training}          & 34                 & 7,675            & 8,186                 \\
\hline 
\textbf{Total}                     & 159                & 14,090           & 16,687                \\
\hline \hline 
\textbf{Testing}                   & 72                 & 5,089            & 6,129                 \\
\hline
\textbf{Complex Testing}           & 12                 & 3,533            & 4,971                 \\
\hline
\textbf{Total}                     & 84                 & 8,622            & 11,000                \\
\hline
\end{tabular}
\caption{Data statistics of the person detection dataset.}
\label{tab:person_detection_statistics}
\end{table}

\begin{table*}[h]
\centering
	\begin{tabular}{| r || c c c | c c c | c c |}
	    \hline
	    \rowcolor{superlightgray} \multicolumn{9}{|c|}{\textbf{Anomaly Dataset -- Train Split}} \\
		\hline
		\multirow{2}{*}{\textbf{Configuration}} & \multicolumn{3}{c|}{\textbf{\# Sequences}} & \multicolumn{3}{c|}{\textbf{\# Frames}} & \multicolumn{2}{c|}{\textbf{Unique Choreographies}}\\
		& Normal & Anomalous & Total & Normal & Anomalous & Total & Normal & Anomalous\\
		\hline \hline
		\textbf{Tilted View} & 285 & 0 & 285 & 185,620 & 0 & 185,620 & 31 & 0 \\
		\hline
		\textbf{Top-down View} & 624 & 0 & 624 & 180,359 & 0 & 180,359 & 19 & 0 \\
		\hline \hline
		\textbf{Total} & 909 & 0 & 909 & 365,979 & 0 & 365,979 & 36 & 0 \\
		\hline
		\multicolumn{7}{c}{\textbf{}} \\
		\hline
	    \rowcolor{superlightgray} \multicolumn{9}{|c|}{\textbf{Anomaly Dataset -- Test Split}} \\
		\hline
		\multirow{2}{*}{\textbf{Configuration}} & \multicolumn{3}{c|}{\textbf{\# Sequences}} & \multicolumn{3}{c|}{\textbf{\# Frames}} & \multicolumn{2}{c|}{\textbf{Unique Choreographies}}\\
		& Normal & Anomalous & Total & Normal & Anomalous & Total & Normal & Anomalous\\
		\hline \hline
		\textbf{Tilted View} & 31 & 151 & 182 & 66,508 & 25,617 & 92,125 & 29 & 20\\
		\hline
		\textbf{Top-down View} & 79 & 418 & 497 & 104,165 & 49,528 & 153,693 & 18 & 12 \\
		\hline \hline
		\textbf{Total} & 110 & 569 & 679 & 170,673 & 75,145 & 245,818 & 34 & 22 \\
		\hline
	\end{tabular}
	\caption{Data statistics of the anomaly dataset's train and test split. The train split does not contain anomalies since the split was made for usage with unsupervised methods. Note that there some choreographies are used in both the tilted- as well as the top-down view, so the total number of unique choreographies is less than the sum from the configurations.}
		\label{tab:anomaly_dataset_statistics}
\end{table*}

\subsection{Data Statistics} \label{sec:statistics}
%\previewcomment{Statistics are the same as given in D3.15}

\tabref{tab:anomaly_dataset_statistics} and \tabref{tab:person_detection_statistics} show the splits of datasets in training and testing data. 
%\subsection{Train and Test Splits} \label{sec:train_and_test_split}
For the case of person detection, there are approximately \num{14}K frames from \num{159} sequences for training and \num{8.6}K frames from \num{84} sequences for testing.
Both the training and testing set have been further split according to the complexity of the scenes. \emph{Scene 2} data were recorded with a camera mounted at different heights, also in this scene the variety of movements is greater than in \emph{Scene 1}, which explains the choice of these sequences for training. In contrast, data from \emph{Scene 1} captured with the camera mounted at \SI{2.50}{\m} and suggested to be used for testing. Complex top-down sequences were captured at \emph{Scene 2} and split to training and testing sets based on the captured person and his activity.

The anomaly dataset consists in total of \num{929} sequences for training and \num{659} sequences for testing. The data splits are designed for usage with unsupervised learning techniques. Therefore the training set only consists of normal sequences. The test set mostly consists of sequences that contain anomalies, but also contains some normal sequences as well. Also including normal sequences in the test split aims at facilitating the evaluation of the false positive rate. Because of the two different camera configurations, used in the recordings in \emph{Scene 1}, both training and test set are split accordingly.

\section{Conclusion}
We presented an extensive dataset of video sequences for monitoring indoor scenes consisting of IR and depth videos captured by a time-of-flight camera of the latest generation. It consists of about \num{1500} sequences for the anomaly detection use case and about \num{240} sequences for the person detection and people counting use case. We described the data and the associated annotations as well as the recording setup and process. The dataset aims at facilitating the development of depth-based algorithms for monitoring indoor spaces in order to allow such functionality to be implemented in a more privacy-preserving way.

\section*{Acknowledgements} This work was partially funded within the \emph{Electronic Components and Systems for European Leadership} (ECSEL) joint undertaking in collaboration with the European Union’s \emph{H2020 Framework Program} and the \emph{Federal Ministry of Education and Research} of the Federal Republic of Germany (BMBF), under grant agreement 16ESE0424 / GA826600 (VIZTA).

Special thanks to the researchers at DFKI and IEE who contributed to the planning and recording of the dataset: Hartmut Feld, Ibrahim Abdelaziz, Dennis Stumpf, Jigyasa Singh Katrolia, Alain Garand, Frederic Garcia, Valeria Serchi, Thomas Solignac. We would also like to thank the students who worked on annotating the data: Sai Srinivas Jeevanandam, Ahmed Elsherif, Brijesh Varsani and Kannan Balakrishnan.

{\small
\bibliographystyle{ieee_fullname}
\bibliography{egbib}
}

\end{document}